\documentclass[11pt]{article}

\usepackage{enumitem}

\usepackage[final]{acl}

\usepackage{times}

\usepackage[T1]{fontenc}

\usepackage[utf8]{inputenc}

\usepackage{microtype}

\usepackage{inconsolata}

\usepackage{graphicx}
\usepackage{subcaption}
\usepackage{float}
\usepackage{algorithm}
\usepackage{algpseudocode}
\usepackage{amsmath}
\usepackage{xcolor}
\usepackage{array}
\usepackage{placeins} 

\newcommand{\methodname}{AGORA}
\newcommand{\methodnamewithspace}{AGORA }

\title{Unifying Language Agent Algorithms with Graph-based Orchestration Engine for Reproducible Agent Research}

\author{
 \textbf{Qianqian Zhang \textsuperscript{1}},
 \textbf{Jiajia Liao \textsuperscript{2}},
 \textbf{Heting Ying \textsuperscript{1}},
 \textbf{Yibo Ma \textsuperscript{1}},
\\
 \textbf{Haozhan Shen \textsuperscript{3}},
 \textbf{Jingcheng Li \textsuperscript{1}},
 \textbf{Peng Liu \textsuperscript{1}},
 \textbf{Lu Zhang \textsuperscript{1}},
\\
 \textbf{Chunxin Fang \textsuperscript{2}},
 \textbf{Kyusong Lee \textsuperscript{1,2}},
 \textbf{Ruochen Xu \textsuperscript{1}},
 \textbf{Tiancheng Zhao\textsuperscript{1,2,*}}
\\
\\
 \textsuperscript{1}Om AI Research,
 \textsuperscript{2}Binjiang Institute of Zhejiang University,
 \\
  \textsuperscript{3}College of Computer Science and Technology, Zhejiang University
\\
  {\tt\small tianchez@zju-bj.com} 
}

\begin{document}
\maketitle
\begin{abstract}
Language agents powered by large language models (LLMs) have demonstrated remarkable capabilities in understanding, reasoning, and executing complex tasks. However, developing robust agents presents significant challenges: substantial engineering overhead, lack of standardized components, and insufficient evaluation frameworks for fair comparison. We introduce Agent Graph-based Orchestration for Reasoning and Assessment (\methodname) \footnote{We made a demo video at: \url{https://www.youtube.com/watch?v=WRH-F1zegKI}. The comparison of agent algorithms across different LLMs is also available at \url{https://huggingface.co/spaces/omlab/open-agent-leaderboard}. Source code of \methodnamewithspace can be found at \url{https://github.com/om-ai-lab/OmAgent}.} , a flexible and extensible framework that addresses these challenges through three key contributions: (1) a modular architecture with a graph-based workflow engine, efficient memory management, and clean component abstraction; (2) a comprehensive suite of reusable agent algorithms implementing state-of-the-art reasoning approaches; and (3) a rigorous evaluation framework enabling systematic comparison across multiple dimensions. Through extensive experiments on mathematical reasoning and multimodal tasks, we evaluate various agent algorithms across different LLMs, revealing important insights about their relative strengths and applicability. Our results demonstrate that while sophisticated reasoning approaches can enhance agent capabilities, simpler methods like Chain-of-Thought often exhibit robust performance with significantly lower computational overhead. \methodnamewithspace not only simplifies language agent development but also establishes a foundation for reproducible agent research through standardized evaluation protocols.
\end{abstract}

\begin{figure*}[ht]
    \centering
    \includegraphics[width=1\textwidth]{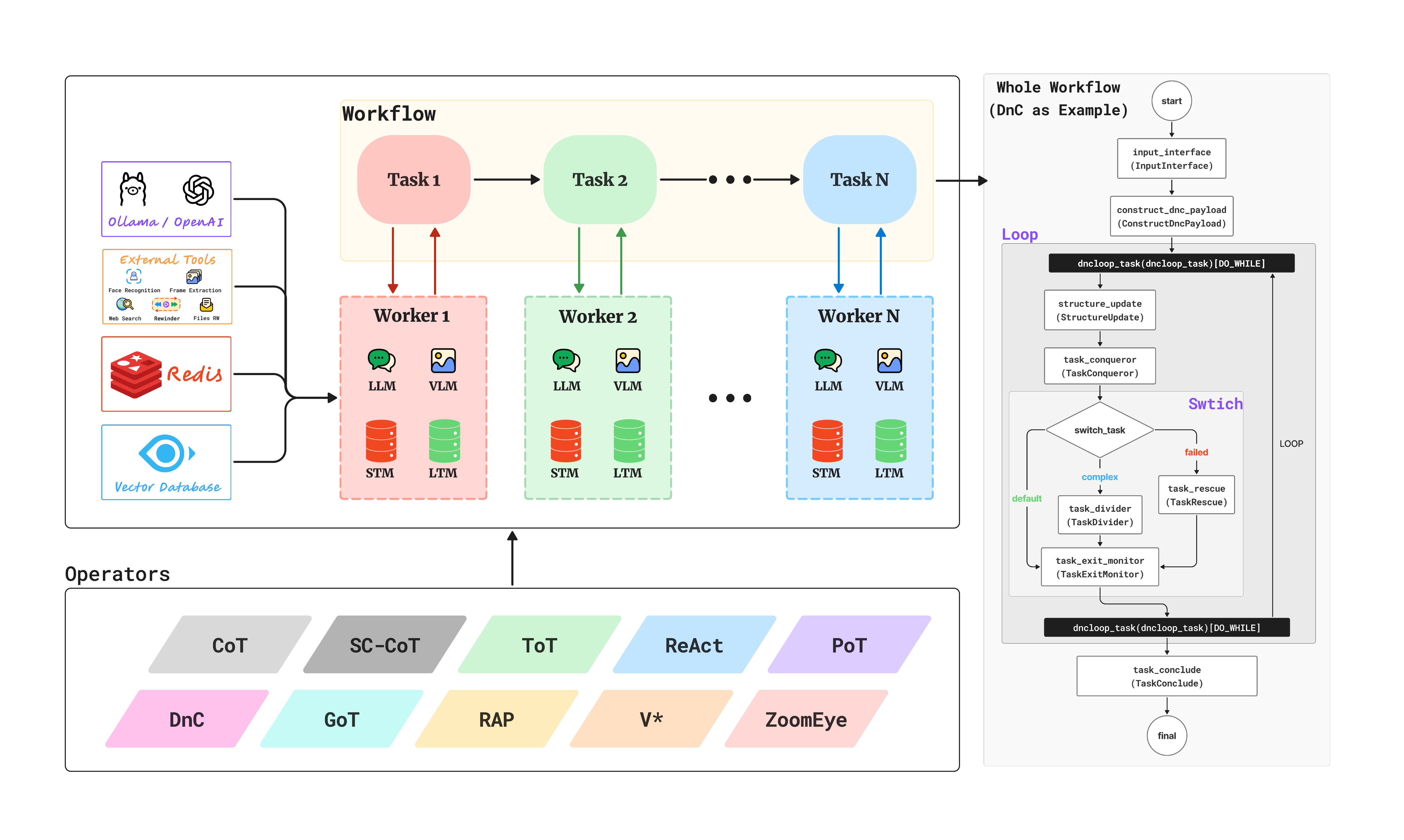}
    \caption{A demonstration of \methodnamewithspace structure.}
    \label{fig:structure}
\end{figure*} 

\section{Introduction}

Language agents powered by large language models (LLMs) are rapidly transforming how we approach complex computational tasks across diverse domains. Industry adoption of these technologies is accelerating, with projections suggesting that 33\% of organizations will implement LLM-based applications by 2025\footnote{\url{https://www.gartner.com/en/articles/intelligent-agent-in-ai?}}. This growing adoption stems from the unprecedented ability of these systems to integrate natural language understanding with action-oriented capabilities.

Despite their promising trajectory, the practical implementation of language agents remains challenging for researchers and developers. Current frameworks often require substantial custom engineering efforts for each application domain, leading to fragmented implementations and difficulty in comparing different approaches. 

To bridge this gap, we present \methodname, a comprehensive framework focused on both practical implementation and scientific evaluation of language agents. \methodnamewithspace provides an integrated environment where researchers can experiment with various reasoning strategies while developers can build robust applications with minimal engineering overhead. Our framework makes three key contributions that differentiate it from existing approaches: a graph-based workflow orchestration engine that simplifies complex task execution; modular agent algorithm support for diverse reasoning paradigms; and easy-to-use client interfaces for evaluation and interaction.

Through systematic evaluation on mathematical and multimodal reasoning tasks, we demonstrate that \methodnamewithspace not only facilitates rapid development but also enables rigorous scientific comparison of different agent paradigms. Our results provide actionable insights for researchers and practitioners navigating the growing landscape of language agent technologies.

\section{Related Work}


Recent years have seen significant development in LLM agent frameworks and evaluation methodologies. Frameworks like LangChain \cite{langchain}, AutoGPT \cite{significant2023autogpt}, and AgentVerse \cite{chen2024agentverse} offer general-purpose infrastructures for agent development, while AutoAgent \cite{tang2025autoagent} provides zero-code solutions through declarative interfaces. Specialized frameworks address domain-specific applications, including ChemCrow \cite{bran2023chemcrow} for chemistry and OS-Copilot \cite{wu2024copilot} for operating systems.
For evaluation, comprehensive benchmark suites such as AgentBench \cite{liu2023agentbench} and WebArena \cite{zhou2023webarena} assess agents across multiple dimensions including reasoning, tool use, and web browsing. Leaderboard platforms like Agent Arena \cite{agent-arena} enable systematic comparison of agents across models, frameworks, and tools through user-driven evaluations. 
A notable benchmark in this space is the Agent Leaderboard \cite{agent-leaderboard}, which primarily evaluates LLMs' tool calling and API interaction capabilities. Our work differs by providing a comprehensive evaluation framework that assesses both the underlying LLM capabilities and the effectiveness of different reasoning language agent algorithms, enabling researchers to understand the interplay between model selection and reasoning strategies.

\section{\methodnamewithspace Framework}

\methodnamewithspace is built on top of the OmAgent framework~\cite{zhang2024omagent}, extending it into a flexible and extensible system for building, orchestrating, and evaluating language agents. It abstracts engineering complexity while exposing essential, reusable components—such as LLMs, VLMs, tools, and workflows—needed to construct powerful and research-friendly agents.

\textbf{Graph-based Workflow Orchestration Engine.}
At the core of \methodnamewithspace is a graph-based orchestration engine designed for modularity and scalability. As shown in Figure~\ref{fig:structure}, the system uses a Directed Acyclic Graph (DAG) where each node represents a task. Tasks are either \textit{simple tasks}—developer-defined custom logic—or \textit{logical tasks}—built-in control flows such as branching and looping. Built on the Conductor library, this engine provides visual representations of workflows, making agent behavior intuitive to trace and debug. It also supports asynchronous, distributed execution, which is ideal for managing long-running, complex agent workflows.

\textbf{Modular Agent Algorithm Support.}
\methodnamewithspace includes a diverse set of agent algorithms such as Chain-of-Thought (CoT), Program-of-Thought (PoT), ReAct, Tree-of-Thought (ToT), and more. Each algorithm is implemented as a modular component, allowing developers to reuse common functions like memory access, LLM inference, or tool use. This structure encourages rapid prototyping, easy extensibility, and consistent evaluation across reasoning paradigms.

\textbf{Client Interfaces for Evaluation and Interaction.}
After constructing an agent, \methodnamewithspace provides a suite of Client interfaces tailored to different usage scenarios. 
\begin{itemize}[itemsep=1pt,topsep=2pt]
\item \textbf{WebPageClient:} delivers a web-based chat interface that allows users to directly interact with the agent in real time, making it particularly suitable for qualitative studies such as usability testing or behavioral observation.
\item \textbf{ProgrammaticClient:} supports automated evaluation using predefined JSON test files, making it ideal for quantitative studies with structured benchmarks—it efficiently runs batch test cases, logs outputs, and summarizes scores. 
\item \textbf{DefaultClient:} offers a lightweight command-line interface, designed for quick testing and debugging of agent logic during development. These clients are plug-and-play and can be easily configured via a configuration file, enabling researchers to seamlessly adapt the interface to different stages of experimentation and evaluation. 
\end{itemize} 
These client interfaces are plug-and-play and can be easily configured via a user-friendly config file, enabling seamless switching based on development or evaluation needs.

\FloatBarrier
\begin{table*}[ht]
    \centering
    \fontsize{8pt}{9pt}\selectfont
  \begin{tabular}{p{2.6cm}p{12.3cm}}
\hline
Agent Algorithms                                                             & Description                                                                                                                                                                                                                                                                                \\
\hline
Chain of Thought (CoT) \cite{wei2022chain}                  & Through encourage reasoning in the prompt, CoT enhances LLMs' reasoning by leverages intermediate steps, improving performance in complex tasks like arithmetic and symbolic reasoning. It can be broadly categorized into two types: Zero-shot-CoT and Few-shot-CoT \cite{NEURIPS2022_8bb0d291}.                                                        
\\ \hline
                                                                                                                             
Self-Consistent CoT (SC-CoT) \cite{wang2022selfconsistency} & SC-CoT extends traditional CoT by generating multiple independent reasoning paths for the same problem and aggregating results through majority voting. This approach addresses the inherent variability in LLM reasoning by exploiting the observation that correct answers tend to emerge more consistently across different reasoning attempts than incorrect ones.                                                                                \\ \hline
Tree of Thoughts (ToT) \cite{yao2023tot}                    &  ToT facilitates advanced decision-making by examining coherent textual units, or "thoughts," as intermediate steps in problem-solving. Unlike traditional token-level approaches, ToT enables LLMs to construct and evaluate a thought tree using methods like Breadth-First Search (BFS) or Depth-First Search (DFS) to derive an optimal chain of thought.                                                                                           
\\ \hline
Reasoning and Acting (ReAct) \cite{yao2022react}            & ReAct allows language models to engage with external environments through an iterative cycle of thought, action, and observation. The model reasons about the current state, executes relevant actions, and processes feedback until it gathers sufficient information to deliver a final response.                                                 
\\ \hline
Program of Thought (PoT) \cite{chen2022pot}                 & PoT is designed to enhance the reasoning capabilities of language models by integrating programming language statements into their outputs. Unlike CoT, PoT leverages the strengths of language models like Codex to generate both text and executable code. 
\\ \hline
Divide-and-Conquer (DnC) \cite{zhang2024omagent}            & DnC enhances problem-solving by decomposing complex issues into manageable sub-problems. In this approach, LLMs alternate between the roles of conqueror, which directly addresses the problem, and divider, which breaks it down into smaller components. The conqueror and the divider operate in an iterative loop until the termination criteria are met.                                                     
\\ \hline                                                                                                                
Graph-of-Thought (GoT) \cite{besta2024graph}                & GoT extends the ToT framework by introducing aggregation and refining transformations, enabling advanced graph-based reasoning. This approach decomposes tasks into identical subtasks, processes them independently, and aggregates sub-responses while leveraging internal loops to refine response quality.
    \\ \hline
Reasoning via Planning (RAP) \cite{hao-etal-2023-reasoning} & RAP enhances LLMs by framing complex reasoning tasks as structured planning problems and employing a Monte Carlo Tree Search (MCTS) framework. The RAP implementation follows a tree-search-based architecture with four main components: selection, expansion, simulation, and backpropagation.  Selection means intelligently choosing promising paths through the reasoning tree; Expansion breaks down complex questions into manageable sub-questions; Simulation evaluates potential solution paths through systematic exploration; and Backpropagation updates the search strategy based on solutions discovered. In contrast to ToT, RAP enables backpropagation in the search framework, enhancing the efficiency of decision-tree traversal. 
\\ \hline
V* \cite{wu2023vguidedvisualsearch}                         & V* introduces a meta-architecture for VLMs, SEAL (Show, sEArch, and TelL), a LLM-guided visual search method that enhances high-resolution image processing through iterative search and contextual reasoning. V* simulates human visual search process and leverages top-down features and contextual guidance to address the limitations of traditional visual encoders. First, V* assesses whether visual search is necessary. If so, the VLM identifies the target object. Subsequently, the LLM-guided search model recursively partitions the image into smaller regions and searches for the target based on the confidence scores derived from contextual cues until the target is located. The information about the identified target is stored in the Visual Working Memory (VWM). Finally, the VLM generates the response using the visual information of all targets stored in the VWM. A implementation of V* is presented in Algorithm \ref{alg:vstar}.  
\\ \hline
ZoomEye \cite{shen2024zoomeyeenhancingmultimodalllms}       & ZoomEye is a training-free agent algorithm that enhances VLM performance on high-resolution images by simulating human zooming behavior. Treating the image as a tree structure, it dynamically explores zoomed-in regions based on visual cues and problem-specific priorities calculated by the VLMs.  \\
\hline    
\end{tabular}

\caption{Agent algorithms implemented in \methodname.}
\label{tab:algorithms}
\end{table*}

\section{Agent Algorithms}

The \methodnamewithspace framework uses modular, reusable components called \textbf{operators} to simplify building and customizing AI systems. Each operator acts as a self-contained unit designed for a specific task, with clear input and output connections that make it easy to integrate into larger workflows.

We implemented various agent algorithms as operators and rigorously evaluated their performance in standardized, controlled environments. A description of the implemented agent algorithms is provided in Table \ref{tab:algorithms}. In particular, RAP enables more reliable and transparent decision-making processes by transforming complex reasoning tasks into systematic planning problems. The RAP implementation follows a tree-search-based architecture with four main components: selection, expansion, simulation, and backpropagation. In contrast to ToT, RAP enables backpropagation in the search framework, enhancing the efficiency of decision-tree traversal.

\subsection{Implemented Agent Algorithms}

In addition, We enhaced ReAct to ReAct-pro inspired by the Reflexion \cite{shinn2023reflexion} implementation. We modified our approach by separating the previously combined Think and Action steps into two distinct model calls, allowing the model to focus more intently on each phase. We also improved PoT by merging short-answer and multiple-choice questions processes into a single workflow consisting of two modules: the program executor and the answer extractor. For GoT, we extend the original GoT implementation into general GoT by allowing it to conduct any tasks other than the predefined tasks like sorting.

\begin{algorithm}[htbp]
\tiny 
\caption{V*}
\label{alg:vstar}
\begin{algorithmic}[1]

\Function{VStar}{Image I, Query T}
    \State VWM $\leftarrow$ Init(I, T)                      
    \State targets $\leftarrow$ LLMIdentify(I, T)          
    \For{each tar in targets}
        \State patchBox $\leftarrow$ getSize(I)
        \While {true}
            \If {patchBox $\leq$ minCropSize} \textbf{break} \EndIf
            \State imagePatch $\leftarrow$ CropImage(I, patchBox)
            \State (scores, subImagePatchs, coords, conf) $\leftarrow$ VisualSearch(imagePatch, tar)          
            \If{conf $\geq$ thresh} \State Store(VWM, tar, coords) \textbf{break} \EndIf
            \State searchQueue $\leftarrow$ \Call{heappush}{priorityQueue, (score, subImagePatch)}
            
            \If {priorityQueue not empty} 
                \State patch $\leftarrow$ \Call{heappop}{priorityQueue}[1] 
                \State PatcheBox $\leftarrow$ getSize(patch)

            \EndIf
        \EndWhile
    \EndFor
    \State \Return LLMAnalyze(VWM)                          
\EndFunction
\end{algorithmic}
\end{algorithm}

\section{Evaluation and Leaderboard}
\subsection{Evaluation Framework}
Our experimental evaluation focused on two distinct domains: unimodal mathematical reasoning tasks and multimodal high-resolution image question-answering reasoning tasks. Mathematical reasoning tasks serve as canonical benchmarks for logical inference and problem decomposition, challenging agents to exhibit systematic reasoning and numerical accuracy. These tasks are inherently language-intensive yet require precise step-by-step deduction, making them ideal for evaluating the core reasoning capabilities of LLMs. Meanwhile, multimodal tasks involving high-resolution image understanding address the growing demand for agents to simulate real-world scenarios where contextual reasoning across diverse inputs is essential.
Comprehensive experiments were conducted across multiple evaluation metrics, agent algorithms, and LLMs to assess reasoning capabilities in both domains.

To evaluate language agents, this study defines four key metrics: accuracy, cost, token usage, and pass rate. Specially, accuracy assesses the proportion of predictions that exactly match the ground-truth response; cost quantifies the total expenditure incurred measured in US dollar. We used API services for close-sourced models and models with more than 70 billion from SiliconFlow\textsuperscript{\footnotemark[1]} and OpenAI\textsuperscript{\footnotemark[2]}; Token usage measures the number of tokens that a language agent uses to generate predictions, and pass rate measures the proportion of valid predictions among all predictions, where a prediction is considered valid if it is neither empty nor null.
\footnotetext[1]{SiliconFlow: https://siliconflow.cn/zh-cn/}
\footnotetext[2]{OpenAI: https://openai.com/}

\subsection{Experimental Setup}
\subsubsection{Mathematical Reasoning Tasks}
The mathematical reasoning benchmarks include:

\textbf{GSM8K} \cite{cobbe2021gsm8k}: A dataset for evaluating language agents' ability to solve elementary math word problems. we conducted the evaluation using 8-shot learning. 

\textbf{AQuA} \cite{ling2017program}: This dataset is specifically designed to reason through diverse algebraic problems to assess reasoning abilities. We employed zero-shot learning in the experiments. 

 \textbf{MATH-500} \cite{hendrycks2021measuring}: A dataset comprising 500 mathematical reasoning problems has been meticulously designed to evaluate the ability of language agents to tackle complex mathematical challenges, where 4-shot learning is applied. 

We applied both commercial and open-source models in the experiments.

   \textbf{Commercial Models:} In our experiment, GPT-3.5 Turbo and GPT-4o from OpenAI, and Doubao-lite-32k from ByteDance were used as LLM for agent algorithms, and GPT-3.5 Turbo was also used for the extraction of AQuA answers.

   \textbf{Open-source models: } We also evaluated open source models like Llama and Qwen for performance and cost effectiveness. We used the following models as the LLMs for Agents: Qwen2.5-72B-Instruct, Qwen2.5-7B-Instruct \cite{qwen2.5}, Qwen2-1.5B-Instruct, Qwen2-0.5B-Instruct \cite{qwen2}, Llama-3.3-70B-Instruct, Llama-3.1-8B-Instruct \cite{llama3modelcard}, InternLM2.5-7B-Chat \cite{cai2024internlm2}, deepseek-r1-1.5B \cite{guo2025deepseek}.

In the experiments, the default setting uses a temperature of 0. More algorithm settings other than default can be found in Appendix \ref{parameter}. 

\subsubsection{Multimodal Reasoning Tasks}
Regarding multimodal reasoning task, we implemented MME-RealWorld \cite{zhang2025mmerealworldmultimodalllmchallenge} as the benchmark. MME-RealWorld aims at solving high-resolution image problems highly relevant to real-world applications. Specifically, we selected images with resolutions between 2K and 4K in the lite version.  We implemented V* and ZoomEye in the evaluation, implementation details can be found in Appendix \ref{parameter}. Because we only applied open source VLMs and  all models used were deployed locally, cost is not involved for evaluation.

\subsection{Mathematical Reasoning Results}


\begin{figure}[ht]
    \centering
    \begin{subfigure}[b]{0.45\textwidth}
        \centering
        \includegraphics[width=\textwidth]{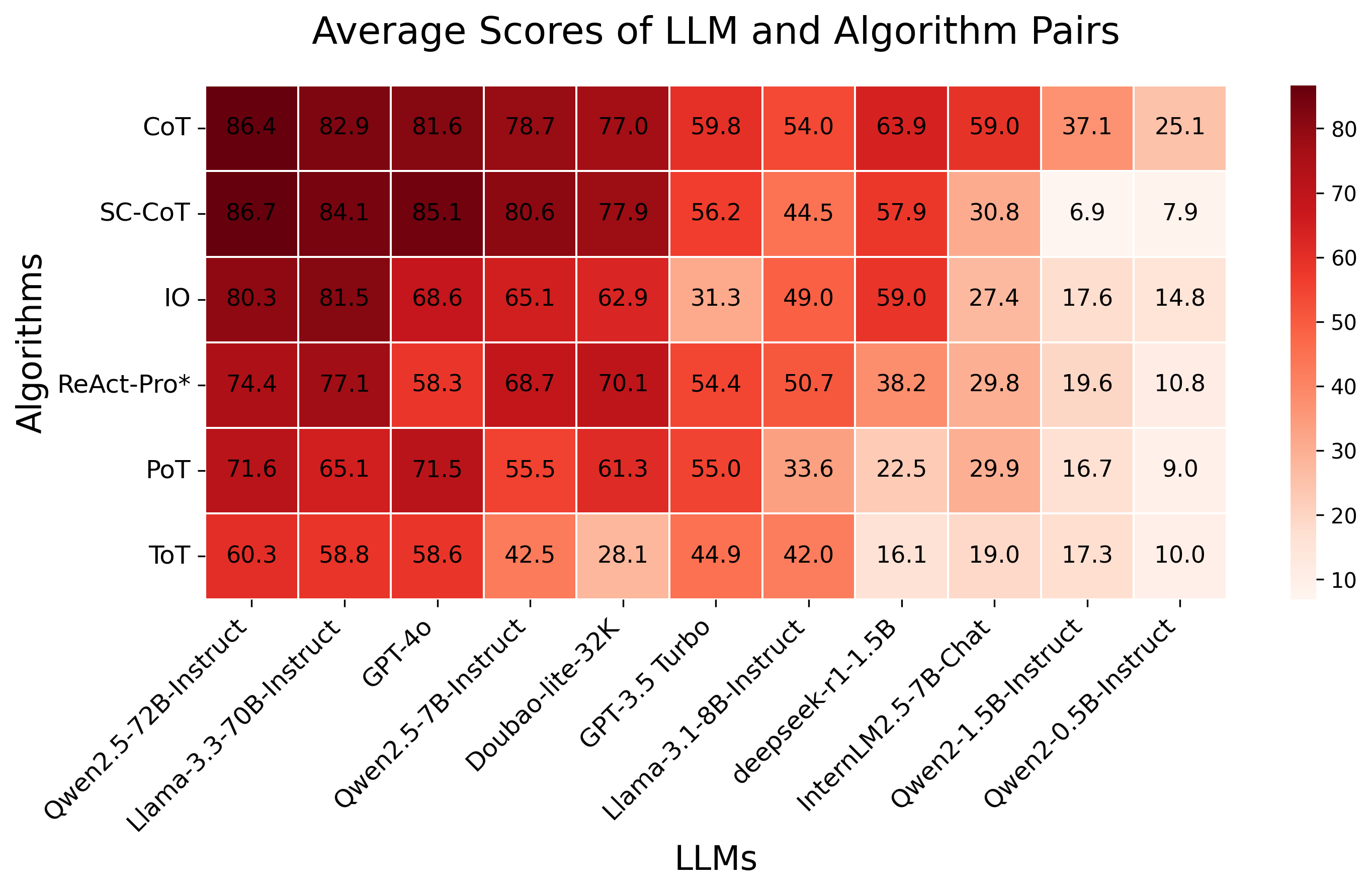}
        \caption{Average scores.}
        \label{fig:fig1}
    \end{subfigure}
    \hfill
    \begin{subfigure}[b]{0.45\textwidth}
        \centering
        \includegraphics[width=\textwidth]{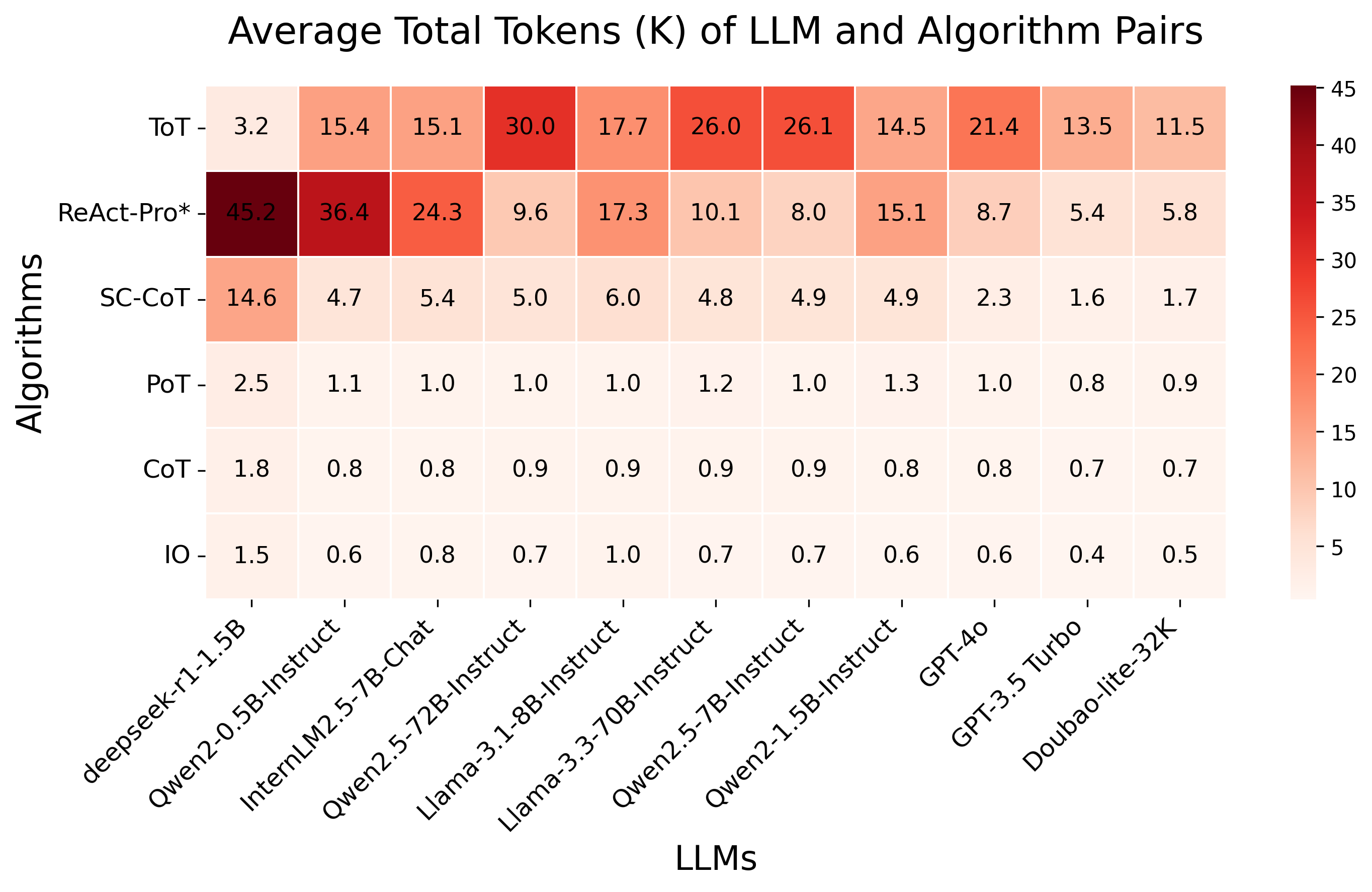}
        \caption{Average input and output token consumptions.}
        \label{fig:fig2}
    \end{subfigure}
    \caption{LLMs and agent algorithms average scores and average token consumptions on mathematical reasoning tasks.}
    \label{fig:matrix}
\end{figure}

\subsubsection{Performance Comparison}


The average scores and average token consumptions of LLM and algorithm pairs are illustrated in Figure \ref{fig:matrix}, where the average token consumption is calculated by first summing the input and output tokens per sample for each dataset, then computing the overall mean across all benchmarks. The comparison details can be found at Open Agent leaderboard \cite{open-agent-leaderboard}. Furthermore, we performed a score versus cost analysis for different LLM agent algorithms, as depicted in Figure \ref{fig:score_vs_cost}. The dashed line in the plot represents an ideal trend line, which serves as a visual benchmark, illustrating the optimal balance between cost and performance. Points on the top-left corner indicate agent-LLM pairs that offer the best possible trade-off between task accuracy and computational cost. Models smaller than 7B parameters were self-hosted locally, thus their cost metrics are not shown. It should be mentioned that GoT, RAP and DnC were excluded from the comparison. GoT is specifically designed to decompose complex tasks into several identical sub-tasks, such as sorting and keyword counting. RAP and DnC was not included due to its high token consumption.


Open-source models with 70 billion parameters have demonstrated exceptional performance compared to other models. Also, Qwen2.5-7B-Instruct surpasses GPT-3.5 Turbo in this task. Surprisingly, deepseek-r1-1.5B, with only 1.5 billion parameters, exhibits remarkable performance by outperforming the InternLM2.5-7B-Chat model. When considering different agent algorithms, the simplest CoT approach also outperforms other agent algorithms while utilizing the least number of tokens.


\subsubsection{Key Findings}
\textbf{Simple agent algorithms show robust performance.}
CoT and SC-CoT algorithm has demonstrated remarkable performance despite their simplicity. Utilizing the Doubao-lite-32k model, CoT achieved an accuracy of 89.31\% on the GSM8K dataset, with a token cost of only \$0.0558. 
 However, SC-CoT encounters challenges with smaller models, which struggle to strictly adhere to instructions, resulting in difficulties parsing the output. 
 Notably, more advanced algorithms, such as PoT and TOT, which incorporate external tools, perform worse on mathematical problems compared to the simpler algorithms. We observed that PoT's reliance on the code generation and parsing capabilities of LLMs does not lead to significant improvements compared to other agent algorithms. In fact, it can have negative effects, particularly with smaller LLM models due to the code generation quality. 
Moreover, the thinking generation and state evaluation for ToT does not significantly reduce the difficulty of reasoning, but rather significantly increases its token usage, which leads to exhibiting poorer performance.

 This phenomenon prompts a reflection on the value of algorithmic complexity. The advantage of simpler methods is primarily reflected in the reduction of error accumulation.
 Complex agent algorithms often involve multiple steps, each potentially introducing errors, whereas a single reasoning chain significantly reduces the risk of error propagation.
CoT's simple prompts are easier to adjust and optimize, making the reasoning process more transparent, easier to understand, and improved. In terms of cost-effectiveness, CoT's advantages are even more apparent. Lower token consumption translates to reduced operational costs, and faster reasoning speeds enhance system responsiveness. Additionally, the straightforward implementation reduces development and maintenance costs.
These findings offer important practical insights. When designing intelligent systems, we should prioritize simple and direct solutions, introducing complexity only when necessary. It is advisable to start with a basic CoT implementation and gradually optimize based on the specific task characteristics, while carefully evaluating the actual benefits of each added complexity.

\textbf{Agent algorithms can be sensitive to prompts.}
We also noticed the importance of prompt design. As shown in Table \ref{tab:react_comparison}, the base ReAct achieved a baseline performance of 34.25\% on the AQuA dataset. Inspired by the Reflexion implementation, we prompt ReAct to ReAct-Pro by separating the previously combined Think and Action steps into two distinct model calls, allowing the model to focus more intently on each phase. This modification alone boosted accuracy to 40.16\%. The real breakthrough came from a remarkably simple addition by including the sentence: "You can take as many steps as needed" in the prompt, we observed an extraordinary increase in accuracy to 64.57\%, an almost 90\% improvement over the baseline. This simple prompt fundamentally transformed the model's behavior patterns. 

\begin{table}[htbp]
    \centering
    \fontsize{8pt}{9pt}\selectfont
    \begin{tabular}{l|l|l|l}
        \hline
        Agent Algorithm & Dataset & LLM & Score \\ \hline
        ReAct & GSM8K & GPT-3.5 Turbo & 38.13 \\ 
        ReAct-Pro & GSM8K & GPT-3.5 Turbo & 74.91 \\ 
        ReAct & AQuA & GPT-3.5 Turbo & 34.25 \\ 
        ReAct-Pro & AQuA & GPT-3.5 Turbo & 64.57 \\  \hline
    \end{tabular}
    \caption{Comparison of ReAct and ReAct-Pro on different datasets.}
    \label{tab:react_comparison}
\end{table}


\textbf{Open-source models are competitive with commercial ones}.
Open-source models at the 70B level, such as Llama-3.3-70B-Instruct and Qwen2.5-72B-Instruct, have shown outputs that exceed those of the closed-source GPT-4o. 
However, the enhancement brought by agent frameworks to top-tier large models (such as GPT and models above 70B) is relatively limited. In some cases, complex agents like ReAct may even lead to a decline in performance. 

\textbf{Small models perform better with simple agent algorithms.}
For smaller models, such as Qwen2.5-7B-Instruct, CoT demonstrates a marked improvement, while PoT shows limited enhancement. This limitation is primarily attributed to the bottleneck in code generation capabilities. 


\subsection{Multimodal Reasoning Results}
\subsubsection{Performance Comparison}
We compared IO, V*, and ZoomEye using various models. The detailed comparison results are shown in Table \ref{tab:multimodal_comparison} in Appendix \ref{mllm_comp}. It is important to note that due to the specific nature of the V* models, we were unable to obtain their token usage data. Overall, the final scores of the same models improved after using the ZoomEye framework, particularly the Qwen2.5-VL-7B-Instruct model, which even outperformed the Qwen2.5-VL-72B-Instruct IO. After applying the agent algorithms, both the input and output token usage increased significantly. Notably, the Qwen2.5-VL models (7B and 72B) demonstrated identical token consumption patterns in IO, which can be attributed to their strong instruction adherence capabilities and the multiple-choice format of the benchmark questions. Moreover, the V* framework received one of the lowest scores, primarily due to its low pass rate.

\subsubsection{Key Findings}

In our experiments, we found that the performance of the models was generally improved after using a multimodal agent workflow like ZoomEye, especially the 7B model outperformed the 72B model. This phenomenon suggests that adopting multimodal agent can effectively provide more visual details in final answer, thus helping the model to generate more accurate answers. Therefore, if computational resources are sufficient, it is recommended to prioritize models with larger parameters to fully leverage their potential. However, if computational resources are limited, smaller models combined with efficient agent workflows can still achieve comparable results.




\section{Discussion}


 In the design of agent systems, it is crucial to prioritize straightforward and direct solutions, incorporating complexity only when necessary. It is recommended to begin with a fundamental CoT that achieves a balance between performance and cost. Complexity can be progressively increased based on task requirements (e.g., using ToT for hierarchical planning when CoT proves insufficient) , ensuring a systematic trade-off between efficiency and task complexity.
For the selection of LLMs, we recommend utilizing models with at least 7 billion parameters or employing reasoning models such as deepseek-r1. This recommendation is primarily due to the tendency of smaller models to exhibit issues with instruction adherence. 
Furthermore, we noticed multimodal agent algorithms like ZoomEye can enhance agent performance by providing valuable visual details. Although larger models should be prioritized when resources allow, smaller models can still yield competitive outcomes.


\section{Conclusions}

In this paper, we present \methodnamewithspace, a comprehensive framework for building and evaluating language agent algorithms that addresses critical challenges of engineering overhead, fragmented implementations, and insufficient evaluation standards.  Our graph-based workflow orchestration engine (built on DAGs) enables dynamic task decomposition and asynchronous distributed execution. Meanwhile, its modular design standardizes agent algorithms (e.g., CoT, V*) for plug-and-play integration. The multi-client evaluation interfaces also facilitate both qualitative user studies and quantitative benchmarking, enabling rigorous cross-algorithm comparisons across LLMs and tasks. We systematically integrated 10 state-of-the-art agent algorithms spanning from CoT to V*, under a unified modular architecture, which reduces engineering overhead.

Our evaluation across mathematical and multimodal tasks revealed several important insights. First, simpler reasoning approaches like CoT often demonstrate robust performance and consume less cost than more complex alternatives. Second, the effectiveness of different agent algorithms varies substantially across different model sizes. Third, for multimodal tasks, specialized agent algorithms like ZoomEye can substantially enhance model performance on high-resolution images, highlighting the value of reasoning strategies using VLMs.

As the field continues to evolve, we believe this framework will serve as a valuable foundation for exploring increasingly sophisticated agent architectures and reasoning approaches. Future work of \methodnamewithspace should focus on: (1) expanding the evaluation framework to encompass broader complex real-world tasks (e.g., tool utilization and web interaction scenarios); (2) developing adaptive agents that dynamically select optimal reasoning strategies based on task characteristics; and (3) prioritizing seamless integration of emerging LLMs via extensions to AGORA’s modular architecture.

\bibliography{acl_latex}

\newpage
\appendix
\section{Agent Algorithm Parameter Settings} \label{parameter}

In the experiments of this paper, the default setting for LLMs uses a temperature of 0. For ReAct-Pro, the parameter is set with a maximum number of steps equal to 10. For SC-CoT, the temperature is 1 and the number of paths is 5; For TOT, we use bfs as the search method, with b as 1, max depth and a max steps are both setted as 6, and the number of evaluations is 3.

The mulitmodal model configration is described as follows:

  \textbf{V*: }The SEAL structure uses specific models trained on llava-7b, including seal\_vqa\_7b and seal\_vsm\_7b. seal\_vqa is responsible for identifying and providing the target objects needed for the search from question, as well as utilizing the data in the VWM(visual working memory) to answer the relevant questions. seal\_vsm combines the common sense knowledge with the context of the image to locate the target object and records its information into VWM. Due to the specificity of the model, parameters such as temperature and max\_tokens were not configured. As for the visual search parameters such as the confidence threshold, we use the same parameters as the original settings: confidence maximum 0.5, minimum 0.3, target cue threshold 6.0, target cue threshold decay 0.7, target cue threshold minimum 3.0. In addition we set 10 as the maximum search steps for each target. The reason for this is that the minimum image size of Vstar is 224×224, which can take an hour or even longer when searching for high-resolution images (e.g., 4K images) if we do not limit the number of search steps.

    \textbf{ZoomEye: }As a more generalized agent visual search framework, we apply and evaluate a variety of mainstream open-source multimodal models, including Llava-v1.5-7B \cite{liu2023visualinstructiontuning}, InternVL2.5-8B \cite{chen2025expandingperformanceboundariesopensource}, Qwen2.5-VL-7B-Instruct \cite{bai2025qwen25vltechnicalreport}, and Qwen2.5-VL-72B-Instruct, which support a wide range of complex multimodal visual questioning tasks. For these VLMs, we set temperature to 0.0 and max\_tokens to 2048. We also set the same parameters as the ZoomEye original settings: 
    
\begin{itemize}
    \item Answering Confidence Threshold:
    \begin{itemize}
        \item Maximum: 0.4
        \item Minimum: 0
    \end{itemize}
    \item Smallest Patch Size: 384
    \item Depth Limit: 5
    \item Number of Intervals: 2
    \item Threshold Decrease: [0.1, 0.1, 0.2]
\end{itemize}

\begin{figure*}[ht]
    \centering
    \includegraphics[width=0.95\textwidth]{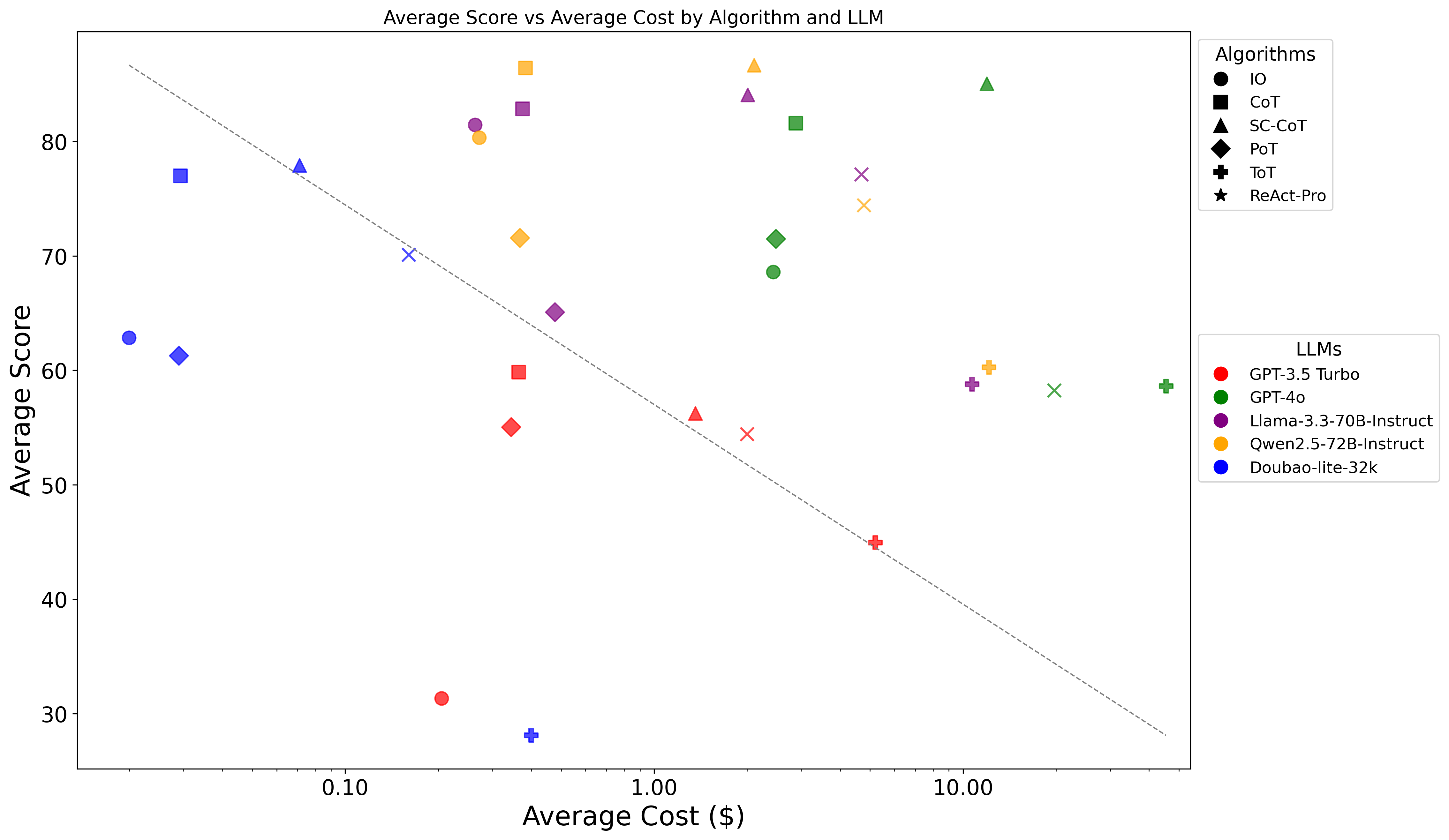}
    \caption{Score versus cost analysis for different LLM agent algorithms. The ideal models appear in the top-left corner with high performance and low cost. Models smaller than 7B parameters were self-hosted locally, thus their cost metrics are not shown.}
    \label{fig:score_vs_cost}
\end{figure*} 
\section{Score Versus Cost Analysis on Mathematical Reasoning}
\label{sec:appendix}

\section{Performance Comparison on Multimodal Reasoning} 
\label{mllm_comp}

\begin{table*}[htbp]
    \centering
    \fontsize{8pt}{9pt}\selectfont
    \setlength{\tabcolsep}{3pt}
    \begin{tabular}{c|l|>{\centering\arraybackslash}p{2cm}|>{\centering\arraybackslash}p{2cm}|>{\centering\arraybackslash}p{2cm}|>{\centering\arraybackslash}p{2cm}|>{\centering\arraybackslash}p{2cm}}
    \hline
    \textbf{Agent} & \textbf{VLMs} & \textbf{Score} & \textbf{Pass Rate} & \textbf{Total Input Tokens} &  \textbf{Total Output Tokens} &  \textbf{All Tokens} \\ 
    \hline
    ZoomEye & Qwen2.5-VL-72B-Instruct & 51.56 & 99.81 & 76,808,965 & 1,276,460 &  78,085,425 \\
    
    ZoomEye & Qwen2.5-VL-7B-Instruct & 48.06 & 96.50 & 94,418,593 & 1,472,836 & 95,891,429 \\
    
    IO & Qwen2.5-VL-72B-Instruct & 44.47 & 100.00 & 6,174,490 & 2,114 & 6,176,604 \\
    
    ZoomEye & InternVL2.5-8B & 43.42 & 99.34 & 153,857,588 & 2,017,170 & 155,874,758 \\
    
    IO & InternVL2.5-8B & 42.95 & 100.00 & 2,779,778 & 2,335 & 2,782,113 \\
    
    IO & Qwen2.5-VL-7B-Instruct & 42.86 & 100.00 & 6,174,490 & 2,114  & 6,176,604 \\
    
    ZoomEye & Llava-v1.5-7B & 31.60 & 98.86 & 113,073,261 & 1,368,724 & 114,441,985 \\
    
    IO & Llava-v1.5-7B & 24.79 & 100.00 & 734,868 & 17,036 & 751,904 \\
    
    V* & seal\_vqa \& seal\_vsm & 15.14 & 72.37 & - & - & - \\
    \hline
    \end{tabular}
    \caption{Performance comparison of different agents and VLMs on MME-RealWorld.}
    \label{tab:multimodal_comparison}
\end{table*}


\end{document}